\documentclass[11pt,a4paper]{article}
\usepackage[nohyperref]{acl2019}
\usepackage[breaklinks]{hyperref}
\usepackage{xcolor}		% make links dark blue
\definecolor{darkblue}{rgb}{0, 0, 0.5}
\hypersetup{colorlinks=true,citecolor=darkblue, linkcolor=darkblue, urlcolor=darkblue}

\usepackage{times}
\usepackage{latexsym}
\usepackage{amsmath}
\usepackage{url}
\usepackage{todonotes} %[disable]

\usepackage{placeins}
\usepackage{tikz}
\usepackage{pgfplots}
\pgfplotsset{width=8cm,compat=1.9}
\usetikzlibrary{pgfplots.dateplot}
\usepackage{pgfplotstable}
\usepackage{filecontents}

\newcommand\model[1]{\textit{#1}} %\textsf
\aclfinalcopy % Uncomment this line for the final submission
\def\aclpaperid{1626} %  Enter the acl Paper ID here

\title{Classification and Clustering of Arguments with\\ Contextualized Word Embeddings}

\author{Nils Reimers, Benjamin Schiller, Tilman Beck,\\ \textbf{Johannes Daxenberger, Christian Stab, Iryna Gurevych} \\
Ubiquitous Knowledge Processing Lab (UKP-TUDA)\\
Department of Computer Science, Technische Universit\"at Darmstadt\\
\url{www.ukp.tu-darmstadt.de}}
\date{}

\begin{document}
\maketitle
\begin{abstract}

We experiment with two recent contextualized word embedding methods (ELMo and BERT) in the context of open-domain argument search. 
For the first time, we show how to leverage the power of contextualized word embeddings to classify and cluster topic-dependent arguments, achieving impressive results on both tasks and across multiple datasets.
For argument classification, we improve the state-of-the-art for the UKP Sentential Argument Mining Corpus by 20.8 percentage points and  for the IBM Debater - Evidence Sentences dataset by 7.4 percentage points. 
For the understudied task of argument clustering, we propose a pre-training step which improves by 7.8 percentage points over strong baselines on a novel dataset, and by 12.3 percentage points for the Argument Facet Similarity (AFS) Corpus.\footnote{Code and models available: \href{https://github.com/UKPLab/acl2019-BERT-argument-classification-and-clustering}{https://github.com/UKPLab/\\acl2019-BERT-argument-classification-and-clustering} }
\end{abstract}

\section{Introduction}

Argument mining methods have been applied to different tasks such as identifying reasoning structures \cite{Stab2014}, assessing the quality of arguments \cite{P17-2039}, or linking arguments from different documents \cite{Cabrio2012NaturalLA}. 
Broadly speaking, existing methods either approach argument mining from the discourse-level perspective (aiming to analyze local argumentation structures), or from an information-seeking perspective (aiming to detect arguments relevant to a predefined topic). 
While discourse-level approaches mostly focus on the analysis of single documents or document collections \cite{eger-daxenberger-gurevych:2017:Long}, information-seeking approaches need to be capable of dealing with heterogeneous sources and topics \cite{shnarch2018} and also face the problem of redundancy, as arguments might be repeated across sources.
As a result, this perspective naturally calls for a subsequent clustering step, which is able to identify and aggregate similar arguments for the same topic. 
In this work, we focus on the latter perspective, referring to it as \emph{open-domain} argument search, and show how contextualized word embeddings can be leveraged to overcome some of the challenges involved in topic-dependent argument classification and clustering.

Identifying arguments for unseen topics is a challenging task for machine learning systems. 
The lexical appearance for two topics, e.g.\ ``net neutrality'' and ``school uniforms'', is vastly different. 
Hence, in order to perform well, systems must develop a deep semantic understanding of both the topic as well as the sources to search for arguments. 
Even more so, clustering similar arguments is a demanding task, as fine-grained semantic nuances may determine whether two arguments (talking about the same topic) are similar.
Figure \ref{fig:clustering_example} gives an example of arguments on the topic ``net neutrality''. 
Both arguments center around the aspect of ``equal access for every Internet user'' but are differently phrased.

\begin{figure}[!htp]
%\footnotesize
\fbox{%
	\begin{minipage}{0.465\textwidth}
	\textbf{A1} The ultimate goal is fast, affordable, open Internet access for everyone, everywhere.
	
    \textbf{A2} If this does not happen, we will create an Internet where only users able to pay for privileged access enjoy the network's full capabilities.
\end{minipage}
}

    \caption{Similar \textit{pro} arguments for the topic ``net neutrality''.}
    \label{fig:clustering_example}
\end{figure}

Contextualized word embeddings, especially ELMo \cite{N18-1202} and BERT \cite{devlin2018bert} could offer a viable solution to this problem.
In contrast to traditional word embeddings like word2vec \cite{word2vec} or GloVe \cite{glove}, these methods compute the embeddings for a sentence on the fly by taking the context of a target word into account. 
This yields word representations that better match the specific sense of the word in a sentence.
In cross-topic scenarios, with which we are dealing in open-domain argument search, contextualized representations need to be able to adapt to new, unseen textual topics.
We thus analyze ELMo and BERT in a cross-topic scenario for the tasks of argument classification and clustering on four different datasets. 
For argument classification, we use the \textit{UKP Sentential Argument Mining Corpus} by \citet{Stab2018b} and the \textit{IBM Debater\textsuperscript{\textregistered}: Evidence Sentences} corpus by \citet{shnarch2018}.
For argument clustering, we introduce a novel corpus on \textit{aspect-based argument clustering} and evaluate the proposed methods on this corpus as well as on the \textit{Argument Facet Similarity  Corpus} \cite{MisraEW16}.

The contributions in this publications are: 
(1)~We frame the problem of open-domain argument search as a combination of topic-dependent argument classification and clustering and discuss how contextualized word embeddings can help to improve these tasks across four different datasets.
(2)~We show that our suggested methods improve the state-of-the-art for argument classification when fine-tuning the models, thus significantly reducing the gap to human performance.
(3)~We introduce a novel corpus on aspect-based argument similarity and demonstrate how contextualized word embeddings help to improve clustering similar arguments in a supervised fashion with little training data.

We present the four different datasets used in this work in Section~\ref{sec:Datasets}, before we discuss our experiments and results on argument classification and clustering in Sections~\ref{sec:ArgumentClassification} and~\ref{sec:ArgumentClustering}.
We conclude our findings for open-domain argument search in Section~\ref{sec:Conclusion}.

\section{Related Work}

In the following, we concentrate on the fundamental tasks involved in open-domain argument search. 
First, we discuss work that experiments with sentence-level argument classification. Second, we review work that provides us with the necessary tools to cluster extracted arguments by their similarity. Third, we take a deeper look into contextualized word embeddings.

\textbf{Argument Classification}, as viewed in this work, aims to identify topic-related, sentence-level arguments from (heterogeneous) documents. \citet{levy2014} identify context-dependent claims (CDCs) by splitting the problem into smaller sub-problems. \citet{rinott2015} extend this work with a pipeline of feature-based models that find and rank supporting evidence from Wikipedia for the CDCs. However, neither of these approaches leverage the potential of word embeddings in capturing semantic relations between words.

\citet{shnarch2018} aim to identify topic-dependent evidence sentences by blending large automatically generated training sets with manually annotated data as initialization step. They use a BiLSTM with GloVe embeddings and integrate the topic via attention. For topic-dependent argument detection, \citet{Stab2018b} deploy a modified LSTM-cell that is able to directly integrate topic information. They show the importance of topic information by outperforming a BiLSTM baseline by around 4.5pp. Yet, their best model only shows mediocre recall for arguments, while showing an even lower precision when compared to their baseline. As argument classification is the first logical step in open-domain argument search, a low performance would eventually propagate further down to the clustering of similar arguments. Hence, in this work, we aim to tackle this problem by leveraging superior contextualized language models to improve on precision and recall of argumentative sentences.

\textbf{Argument Clustering} aims to identify similar arguments. Previous research in this area mainly used feature-based approaches in combination with traditional word embeddings like word2vec or GloVe. 
\newcite{boltuvzic2015} applied hierarchical clustering on semantic similarities between users' posts from a two-side online debate forum using word2vec. 
\newcite{P18-1023} experimented with different word embeddings techniques for (counter)argument similarity.
\newcite{MisraEW16} presented a new corpus on argument similarity on three topics. They trained a Support Vector Regression model using different hand-engineered features including custom trained word2vec. \newcite{trabelsi2015extraction} used an augmented LDA to automatically extract coherent words and phrases describing arguing expressions and apply constrained clustering to group similar viewpoints of topics.

In contrast to previous work, we apply argument clustering on a dataset containing both relevant and non-relevant arguments for a large number of different topics which is closer to a more realistic setup.

\textbf{Contextualized word embeddings} compute a representation for a target word  based on the specific context the word is used within a sentence. In contrast, traditional word embedding methods, like  word2vec or GloVe, words are always mapped to the same vector.
Contextualized word embeddings tackle the issue that words can have different senses based on the context. Two approaches that became especially popular are ELMo \cite{N18-1202} and BERT \cite{devlin2018bert}.

ELMo (Embeddings from Language Models) representations are derived from a bidirectional language model, that is trained on a large corpus. Peters et al.\ combine a character-based CNN with two bidirectional LSTM layers. The ELMo representation is then derived from all three layers.

BERT (Bidirectional Encoder Representations from Transformers) uses a deep transformer network \cite{Attention_is_all_you_need} with 12 or 24 layers to derive word representations. Devlin et al.\ presented two new pre-training objectives: the ``masked language model'' and the ``next sentence prediction'' objectives. They demonstrate that the pre-trained BERT models can be fine-tuned for various tasks, including sentence classification and sentence-pair classification. 

ELMo and BERT were primarily evaluated on datasets where the test and training sets have comparable distributions. In cross-topic setups, however, the distributions for training and testing are vastly different. It is unclear, whether ELMo and BERT will be able to adapt to this additional challenge for cross-topic argument mining.

\section{Datasets}
\label{sec:Datasets}
No dataset is available that allows evaluating open-domain argument search end-to-end. 
Hence, we analyze and evaluate the involved steps (argument classification and clustering) independently. 

\subsection{Argument Classification}

To our knowledge, to date there are only two suitable corpora for the task of topic-dependent argument classification. 

\textbf{UKP Corpus.} The \textit{UKP Sentential Argument Mining Corpus} by \citet{Stab2018b} (henceforth: \textit{UKP corpus}) annotated 400 documents with 25,492 sentences on eight controversial topics with the labels: \textit{pro}/\textit{con}/\textit{no} \textit{argument}.

\textbf{IBM Corpus.} The \textit{IBM Debater\textsuperscript{\textregistered}: Evidence Sentences} by \citet{shnarch2018} (henceforth: \textit{IBM corpus}) contains 118 topics drawn from different debate portals. For each topic, \citet{shnarch2018} extracted sentences from Wikidata that were in turn annotated by crowd-workers (10 for each topic-sentence pair) with one of the two labels: \textit{evidence} or \textit{no evidence} in regard to the topic. 

\subsection{Argument Clustering}\label{sec_argument_similarity_corpus}
Topic-dependent argument clustering is an understudied problem with few resources available. Arguments on controversial topics usually address a limited set of aspects, for example, many arguments on ``nuclear energy'' address safety concerns. Argument pairs addressing the same aspect should be assigned a high similarity score, and arguments on different aspects a low score. To date, the only available resource of that kind we are aware of, is the Argument Facet Similarity (AFS) Corpus \cite{MisraEW16}.

\textbf{AFS Corpus.} The AFS corpus annotates similarities of arguments pairwise.
\newcite{MisraEW16} aimed to create automatic summaries for controversial topics. 
As an intermediate step, they extracted 6,000 sentential argument pairs from curated online debating platforms for three topics and annotated them on a scale from 0 (``different topic'') to 5 (``completely equivalent''). A drawback of this corpus is that the arguments are curated, i.e., the dataset does not include noise or non-relevant arguments. Furthermore, the corpus covers only three different topics.

\textbf{UKP ASPECT Corpus.} To remedy these shortcomings, we create a new corpus with annotations on similar and dissimilar sentence-level arguments \cite{Stab2018b}, referred to as the Argument Aspect Similarity (UKP ASPECT) Corpus in the following.\footnote{The dataset is available at \href{http://www.ukp.tu-darmstadt.de/data}{http://www.ukp.tu-darmstadt.de/data}}
 The UKP ASPECT corpus consists of sentences which have been identified as arguments for given topics using the ArgumenText system \cite{Stab2018a}. 
 The ArgumenText system expects as input an arbitrary topic (query) and searches a large web crawl for relevant documents. Finally, it classifies all sentences contained in the most relevant documents for a given query into \textit{pro}, \textit{con} or non-arguments (with regard to the given topic).

We picked 28 topics related to currently discussed issues from technology and society. To balance the selection of argument pairs with regard to their similarity, we applied a weak supervision approach.
For each of our 28 topics, we applied a sampling strategy that picks randomly two \textit{pro} or \textit{con} argument sentences at random, calculates their similarity using the system by \newcite{MisraEW16}, and keeps pairs with a probability aiming to balance diversity across the entire similarity scale. This was repeated until we reached 3,595 arguments pairs, about 130 pairs for each topic.

The argument pairs were annotated on a range of three degrees of similarity (no, some, and high similarity) with the help of crowd workers on the Amazon Mechanical Turk platform. To account for unrelated pairs due to the sampling process, crowd workers could choose a fourth option.\footnote{The exact layout of the Human Intelligence Task (HIT) guidelines, as well as agreement statistics can be seen in the appendix.} 
We collected seven assignments per pair and used Multi-Annotator Competence Estimation (MACE) with a threshold of 1.0 \cite{hovy2013learning} to consolidate votes into a gold standard. 
About 48\% of the gold standard pairs are labeled with \textit{no similarity}, whereas about 23\% resp.\ 13\% are labeled with \textit{some} resp.\ \textit{high similarity}. Furthermore, 16\% of the pairs were labeled as containing invalid argument(s) (e.g. irrelevant to the topic at hand).

We asked six experts (graduate research staff familiar with argument mining) to annotate a random subset of 50 pairs from 10 topics. The resulting agreement among experts was Krippendorff's $\alpha$ = 0.43 (binary distance) resp. 0.47 (weighted distance\footnote{Reduced distance of 0.5 between \textit{high} and \textit{some similarity}, otherwise 1.}), reflecting the high difficulty of the task. Krippendorff's $\alpha$ agreement between experts and the gold standard from crowd workers was determined as 0.54 (binary) resp. 0.55 (weighted distance).

\section{Argument Classification}
\label{sec:ArgumentClassification}
As a first task in our pipeline of open-domain argument search, we focus on topic-dependent, sentence-level argument classification. To prevent the propagation of errors to the subsequent task of argument clustering, it is paramount to reach a high performance in this step.

\subsection{Experimental Setup}
For the UKP Corpus, we use the proposed evaluation scheme by \citet{Stab2018b}: The models are trained on the train split (70\% of the data) of seven topics, tuned on the dev split (10\%) of these seven topics, and then evaluated on the test split (20\%) of the eighth topic. A macro $F_1$-score is computed for the 3-label classes and scores are averaged over all topics and over ten random seeds. 
For the IBM Corpus, we use the setup by \citet{shnarch2018}: Training on 83 topics (4,066 sentences) and testing on 35 topics (1,719 sentences). We train for five different random seeds and report the average accuracy over all runs.

\subsection{Methods}
We experiment with a number of different models and distinguish between models which use topic information and ones that do not.

\textbf{bilstm.} This model was presented as a baseline by \citet{Stab2018b}. It trains a bi-directional LSTM network on the sentence, followed by a softmax classifier and has no information about the topic. As input, pre-trained word2vec embeddings (Google News dataset) were used.

\textbf{biclstm.} \citet{Stab2018b} presented the contextualized LSTM (clstm), which adds topic information to the \textit{i-} and \textit{c-}cells of the LSTM. The topic information is represented by using pre-trained word2vec embeddings.

\textbf{IBM.}  \citet{shnarch2018} blend large automatically generated training sets with manually annotated data in the initialization step. They use an LSTM with 300-d GloVe embeddings and integrate the topic via attention. We re-implemented their system, as no official code is available.  

We experiment with these three models by replacing the word2vec / GloVe embeddings with ELMo and BERT embeddings. The ELMo embeddings are obtained by averaging the output of the three layers from the pre-trained 5.5B ELMo model. For each token in a sentence, we generate  a BERT embedding with the pre-trained BERT-large-uncased model.

Further, we evaluate fine-tuning the transformer network from BERT for our datasets:

\textbf{BERT.} We add a softmax layer to the output of the first token from BERT and fine-tune the network for three epochs with a batch size of 16 and a learning rate of 2e-5. We only present the sentence to the BERT model.

\textbf{BERT$_\text{topic}$.} We add topic information to the BERT network by changing the input to the network. We concatenate the topic and the sentence (separated by a special \texttt{[SEP]}-token) and fine-tune the network as mentioned before.

\subsection{Results and Analysis}
In the following, we present and analyze the results.

\begin{table*}[t]
\centering 
\footnotesize
%\resizebox{0.75\textwidth}{!}{%
\begin{tabular}{|l|c|c|c|c|c||c|}
\hline
\textbf{Model} & \multicolumn{5}{c||}{\textbf{UKP Corpus}} & \textbf{IBM} \\ \hline
 & \textbf{F$_1$} & \textbf{P$_\text{arg+}$} & \textbf{P$_\text{arg-}$} & \textbf{R$_\text{arg+}$} & \textbf{R$_\text{arg-}$} & \textbf{Accuracy} \\ \hline
\multicolumn{7}{|l|}{\textbf{Without topic information}}  \\ \hline
\model{bilstm} \cite{Stab2018b} & .3796 & .3484  & .4710  & .0963 & .2181 & .7201  \\
\model{bilstm$_\text{ELMo}$} & .4382  & .4639  & .5088  & .1840  & .2778 & .7574 \\
\model{bilstm$_\text{BERT}$} & .4631  & .5051  & .5079  & .2074  & .3076 & .7476 \\
\model{BERT-base} & .4680  & .5521  & .5397  & .2352  & .2800 & .7928 \\
\model{BERT-large} & .5019  & \textbf{.5844}  & .5818  & .2917  & .3154 & .8021 \\ \hline 
\multicolumn{7}{|l|}{\textbf{With topic information}}  \\ \hline
\model{outer-att} \cite{Stab2018b} & .3873 & .3651 & .4696 & .1042  & .2381 & -  \\
\model{biclstm} \cite{Stab2018b} & .4242  & .2675  & .3887  & .2817  & .4028  & - \\ 
\model{biclstm$_\text{ELMo}$} & .3924  & .2372  & .4381  & .0317  & .3955 & -  \\
\model{biclstm$_\text{BERT}$} & .4243  & .3431  & .4397  & .1060  & .4275 & - \\ \hline
\model{IBM} \cite{shnarch2018} & - & - & - & - & - & $\sim$ .74  \\
\model{IBM (reproduced)} & - & - & - & - & - &  .7288  \\
\model{IBM$_\text{ELMo}$} & - & - & - & - & - &  .7651  \\
\model{IBM$_\text{BERT}$} & - & - & - & - & - &  .7480  \\ \hline
\model{BERT-base$_\text{topic}$} & .6128  & .5048  & .5313  & .4698  & \textbf{.5795} & \textbf{.8137}  \\
\model{BERT-large$_\text{topic}$} & \textbf{.6325}  & .5535  & \textbf{.5843}  & \textbf{.5051}  & .5594 & .8131 \\ \hline
Human Performance & .8100 & - & - & - & - & -\\ \hline
\end{tabular}
%}
\caption{Results of each model for sentence-level argument classification using cross-topic evaluation on the UKP Sentential Argument Mining Corpus and on the IBM Debater\textsuperscript{\textregistered} - Evidence Sentences dataset. Blank fields result from dataset-specific models. \textbf{P}: precision, \textbf{R}: recall, \textbf{arg+}: \textit{pro}-arguments, \textbf{arg-}: \textit{con}-arguments.}
\label{table_argument_detection}
\end{table*}

\textbf{UKP Corpus.} Replacing traditional embeddings in the \model{bilstm} by contextualized word embeddings improves the model's performance by around 6pp and 8pp in F$_1$ for ELMo and BERT (see Table \ref{table_argument_detection}). The fine-tuned \model{BERT-large} improves by even 12pp over the baseline \model{bilstm} and by this also outperforms \model{bilstm$_\text{BERT}$} by around 4pp. Hence, using an intermediary BiLSTM layer for the \model{BERT} model even hurts the performance. 

Using ELMo and BERT embeddings in the topic-integrating \model{biclstm} model significantly decreases the performance, as compared to their performance in the \model{bilstm}. The contextualized word embedding for a topic is different to the one of a topic appearing in a sentence and the \model{biclstm} fails to learn a connection between them.   

Including the topic into the fine-tuned BERT models increases the F$_1$ score by approx.\ 14.5pp and 13pp for \model{BERT-base} and \model{BERT-large}. This is due to a vast increase in recall for both models; while changes in precision are mostly small, recall for positive and negative arguments increases by at least 21pp for both models. As such, \model{BERT-large$_\text{topic}$} also beats the \model{biclstm} by almost 21pp in F$_1$ score and represents a new state-of-the-art on this dataset. 

While the gap to human performance remains at around 18pp in F$_1$, our proposed approach decreases this gap significantly as compared to the previous state-of-the-art. 
Based on preliminary experimental results, we suspect that this gap can be further reduced by adding more topics to the training data.

The results show that (1) the \model{BERT-[base/large]} models largely improve F$_1$ and precision for arguments and (2) leveraging topic-information yields another strong improvement on the recall of argumentative sentences. The usefulness of topic-information has already been shown by \citet{Stab2018b} through their \model{biclstm} and stems from a much higher recall of arguments while losing some of the precision when compared to their \model{bilstm}. Yet, their approach cannot hold to BERT's superior architecture; the topic-integrating BERT models \model{BERT-base$_\text{topic}$} and \model{BERT-large$_\text{topic}$} not only compensate for the \model{biclstm}'s drop in precision, but also increase the recall for \textit{pro} and \textit{con} arguments by at least 18pp and 15pp. We account this performance increase to BERT's multi-head attention between all word pairs, where every word in a sentence has an attention value with the topic (words).

\textbf{IBM corpus.} As a baseline for models that do not use any topic information, we train three simple BiLSTMs with ELMo, BERT, and 300-d GloVe embeddings and compare them to the fine-tuned base and large BERT models. As Table \ref{table_argument_detection} shows, BERT and ELMo embeddings perform around 2.7 and 3.7pp better in accuracy than the GloVe embeddings. \model{BERT-base} yields even 7pp higher accuracy, while its difference to the large model is only +1pp.

Both \model{BERT-base} and \model{BERT-large} outperform the baseline \model{IBM} set by \citet{shnarch2018} already by more than 6pp in accuracy\footnote{Please note that we refer to our reproduced baseline. Also, the original baseline's performance by \citet{shnarch2018} can only be guessed, since the numbers are drawn from a figure and do not appear in the text.}. The topic integrating models \model{IBM$_{ELMo}$} and \model{IBM$_{BERT}$} do not improve much over their BiLSTM counterparts, which do not use any topic information. Similar to the conclusion for the UKP corpus, we attribute this to the different embedding vectors we retrieve for a topic as compared to the vectors for a topic mention within a sentence. \model{BERT-base$_\text{topic}$} and \model{BERT-large$_\text{topic}$} show the largest improvement with 8pp over the baseline and represent a new state-of-the-art on this dataset.
The fine-tuned BERT models show vast improvements over the baseline, which is on par with the findings for the UKP corpus. 

Yet, in contrast to the results on the UKP corpus, adding topic information to the fine-tuned BERT models has only a small effect on the score. This can be explained with the different composition of both corpora: while sentences in the UKP corpus may only be implicitly connected to their related topic (only 20\% of all sentences contain their related topic), sentences in IBM's corpus all contain their related topic and are thus explicitly connected to it (although topics are masked with a placeholder). Hence, in the IBM corpus, there is much less need for the additional topic information in order to recognize the relatedness to a sentence.

\section{Argument Clustering}
\label{sec:ArgumentClustering}

Having identified a large amount of argumentative text for a topic, we next aim at grouping the arguments talking about the same aspects.

For any clustering algorithm, a meaningful similarity between argument pairs is crucial and needs to account for the challenges regarding argument aspects, e.g., different aspect granularities, context-dependency or aspect multiplicity. Another requirement is the robustness for topic-dependent differences.

Therefore, in this section, we study how sentence-level argument similarity and clustering can be improved by using contextualized word embeddings. We evaluate our methods on the \textit{UKP ASPECT} and the AFS corpus (see Section~\ref{sec_argument_similarity_corpus}).

\begin{table*}[t]
\centering 
\footnotesize
\begin{tabular}{|l|c|c|c|c|c|c|}
\hline
& \multicolumn{3}{c}{\textbf{Without Clustering}} & \multicolumn{3}{|c|}{\textbf{With Clustering}}  \\ \hline
\textbf{Model} & \textbf{F$_\text{mean}$} & \textbf{F$_\text{sim}$} & \textbf{F$_\text{dissim}$} & \textbf{F$_\text{mean}$} & \textbf{F$_\text{sim}$} & \textbf{F$_\text{dissim}$}  \\ \hline
Human Performance & .7834 & .7474 & .8194 & .7070 & .6188 & .7951 \\ 
Random predictions & .4801 & .3431 & .6171 & .4253 & .3162 & .5344 \\ \hline
\multicolumn{7}{|l|}{\textbf{Unsupervised Methods}} \\ \hline
\model{Tf-Idf} & .6118 & .5230 & .7007 & .5800 & .4892 & .6708 \\
\model{InferSent - fastText} &  .6621 & .5866 & .7376 & .6344 & .5443 & .7584 \\
\model{InferSent - GloVe} & .6494 & .5472 & .7517 & .6149 & .4587 & .7711 \\
\model{GloVe Embeddings} & .6468 & .5632 & .7304 & .5926 & .4605 & .7246 \\
\model{ELMo Embeddings} & .6447 & .5355 & .7538 & .6366 & .5347 & .7384 \\
\model{BERT Embeddings} & .6539 & .5232 & .7848 & .6070 & .4818 & .7323 \\ \hline
\multicolumn{7}{|l|}{\textbf{Supervised Methods: Cross-Topic Evaluation}} \\ \hline
\model{BERT-base} & \textbf{.7401} & \textbf{.6695} & .8107 & .7007 & \textbf{.6269} & .7746 \\ 
\model{BERT-large} & .7244 & .6297 & \textbf{.8191} & \textbf{.7135} & .6125 & \textbf{.8146} \\ \hline
\end{tabular}
\caption{$F_1$ scores on the UKP ASPECT Corpus.}
\label{table_argument_clustering}
\end{table*}

\subsection{Clustering Method}

We use agglomerative hierarchical clustering \cite{Day1984} to cluster arguments.

We use the average linkage criterion to compute the similarity between two cluster $A$ and $B$:
$\frac{1}{|A||B|} \sum_{a \in A} \sum_{b \in B} d(a,b)$,
for a given similarity metric $d$. As it is a priori unknown how many different aspects are discussed for a topic (number of clusters), we apply a stopping threshold which is determined on the train set.

We also tested the k-means and the DBSCAN clustering algorithms, but we found that agglomerative  clustering generally yielded better performances in preliminary experiments.

Agglomerative clustering uses a pairwise similarity metric $d$ between arguments. We propose and evaluate various similarity metrics in two setups: (1) Without performing a clustering, i.e.\ the quality of the metric is directly evaluated (\textit{without clustering} setup), and (2) in combination with the described  agglomerative clustering method (\textit{with clustering} setup).

\subsection{Experimental Setup}
We differentiate between unsupervised and supervised methods. Our unsupervised methods include no pre-training whereas the supervised methods use some data for fine-tuning the  model.
For the UKP ASPECT corpus, we binarize the four labels to only indicate similar and dissimilar argument pairs. Pairs labeled with \textit{some} and \textit{high similarity} were labeled as \textit{similar}, pairs with \textit{no similarity} and \textit{different topic} as \textit{dissimilar}. 

We evaluate methods in a 4-fold cross-validation setup: seven topics are used for testing and 21 topics are used for fine-tuning. Final evaluation results are the average over the four folds. In case of supervised clustering methods, we use 17 topics for training and four topics for tuning.
In their experiments on the AFS corpus, \citet{MisraEW16} only performed a within-topic evaluation by using 10-fold cross-validation. 
As we are primarily interested in cross-topic performances, we evaluate our methods also cross-topic: we train on two topics, and evaluate on the third.

\subsection{Evaluation}
\label{ssec:evaluation}

For the UKP ASPECT dataset we compute the marco-average F$_\text{mean}$ for the $F_1$-scores for the \texttt{similar}-label (F$_\text{sim}$) and for the \texttt{dissimilar}-label  (F$_\text{dissim}$). 

In the \textit{without clustering} setup, we compute the similarity metric ($d(a,b)$) for an argument pair directly, and assign the label \texttt{similar} if it exceeds a threshold, otherwise \texttt{dissimilar}. The threshold is determined on the train set of a fold for unsupervised methods. For supervised methods, we use a held-out dev set.

In the \textit{with clustering} setup, we use the similarity metric to perform  agglomerative clustering. This assigns each argument exactly one cluster ID. Arguments pairs in the same cluster are assigned the label \texttt{similar}, and argument pairs in different clusters are assigned the label \texttt{dissimilar}. We use these labels to compute F$_\text{sim}$ and F$_\text{dissim}$ given our gold label annotations. 

For the AFS dataset, \newcite{MisraEW16} computed the correlation between the predicted similarity and the annotated similarity score. They do not mention which correlation method they used. In our evaluation, we show Pearson correlation ($r)$ and Spearman's rank correlation coefficient ($\rho$).

\subsection{Similarity Metrics}
We experiment with the following methods to compute the similarity between two arguments.

\textbf{Tf-Idf.} We computed the most common words (without stop-words) in our training corpus and compute the cosine similarity between the Tf-Idf vectors of a sentence. 

\textbf{InferSent.} We compute the cosine-similarity between the sentence embeddings returned by InferSent \cite{conneauInferSent}. 

\textbf{Average Word Embeddings.} We compute the cosine-similarity between the average word embeddings for GloVe, ELMo and BERT.

\textbf{BERT.} We fine-tune the BERT-uncased model to predict the similarity between two given arguments. We add a sigmoid layer to the special \textit{[CLS]} token and trained it on some of the topics. We fine-tuned for three epochs, with a learning rate of 2e-5 and a batch-size of 32.

\textbf{Human Performance.} We approximated the human upper bound on the UKP ASPECT corpus in the following way: we randomly split the seven pair-wise annotations in two groups, computed their corresponding MACE \cite{hovy2013learning} scores and calculated F$_\text{sim}$, F$_\text{dissim}$ and F$_\text{mean}$. We repeated this process ten times and averaged over all runs (\textit{without clustering} setup). For the \textit{with clustering} setup, we applied agglomerative hierarchical clustering on the MACE scores of one of both groups and computed the evaluation metrics using the other group as the gold label. 
For the AFS dataset, \citet{MisraEW16} computed the correlation between the three human annotators.

\begin{table}[t]
\centering 
\footnotesize
\begin{tabular}{|l|c|c|}
\hline
& \multicolumn{2}{c|}{\textbf{Average}}  \\ \cline{2-3}
& $r$ & $\rho$ \\ \hline
 Human Performance & .6767 & - \\
\hline
\multicolumn{3}{|l|}{\textbf{Unsupervised Methods}} \\
\hline
\model{Tf-Idf} & .4677 & .4298 \\
\model{InferSent - fastText} & .2519 & .2423  \\
\model{InferSent - GloVe} & .2708 & .2663  \\
\model{GloVe Embeddings} & .3240 & .3400 \\
\model{ELMo Embeddings} & .2827 & .2675 \\
\model{BERT Embeddings} & .3539 & .3507 \\ \hline
\multicolumn{3}{|l|}{\textbf{Supervised Methods: Within-Topic Evaluation}}  \\ \hline
\model{SVR} \cite{MisraEW16}  &  .6333 & - \\ 
\model{BERT-base}  & .7475 & .7318 \\ 
\model{BERT-large} & .7256 & .6959 \\ \hline 
\multicolumn{3}{|l|}{\textbf{Supervised Methods: Cross-Topic Evaluation}}  \\ \hline
\model{BERT-base} & .5849 &  .5723 \\ 
\model{BERT-large} & .6202 & .6034 \\ 
\hline
\end{tabular}
\caption{Pearson correlation $r$ and Spearman's rank correlation $\rho$ on the AFS dataset \cite{MisraEW16} averaged over the three topics. 
}
\label{tab:afs_corpus_small}
\end{table}

\subsection{Results and Analysis}
\textbf{Unsupervised Methods.} Table \ref{table_argument_clustering} shows the performance on the novel UKP ASPECT Corpus. When evaluating the argument similarity metrics directly (\textit{without clustering} setup), we notice no large differences between averaging GloVe, ELMo or BERT embeddings. These three setups perform worse than applying InferSent with fastText embeddings. \model{Tf-Idf} shows the worst performance.
In Table \ref{tab:afs_corpus_small}, we show the performances for the AFS corpus (detailed results in the appendix, Table \ref{tab:afs_corpus_full}). In contrast to the ASPECT Corpus, the \model{Tf-Idf} method achieves the best performance and \model{InferSent - fastText} embeddings achieved the worst performance. As for the ASPECT Corpus, ELMo and BERT embeddings do not lead to an improvement compared to averaged GloVe embeddings.

Unsupervised methods compute some type of similarity between sentence pairs. However, as our experiments shows, this similarity notion is not necessarily the notion needed for the task.

\textbf{Supervised Methods.} We fine-tune the BERT model for some of the topics and study the performance on unseen topics. For the ASPECT Corpus, we observe a performance increase of 7.8pp. Identifying dissimilar arguments (F\textsubscript{dissim}) is on-par with the human performance, and identifying similar arguments achieves an F-score of .67, compared to .75 for human annotators. For the AFS dataset, we observe that fine-tuning the BERT model significantly improves the performance by 11pp compared to the previous state-of-the-art from \citet{MisraEW16}.

In a cross-topic evaluation setup on the AFS dataset, we observe that the performance drops to .57 Spearman correlation. This is still significantly larger than the best unsupervised method.

We evaluated the effect of the training set size on the performance of the BERT model for the ASPECT Corpus. A certain number of topics were randomly sampled and the performance was evaluated on distinct topics. This process was repeated 10 times with different random seeds \cite{single_score_evaluation}. Table \ref{table_clustering_small_train_data} shows the averaged results.

By allowing fine-tuning on five topics we are able to improve the F\textsubscript{mean}-score to .71 compared to .65 when using BERT without fine-tuning (\textit{without clustering} setup). Adding more topics then slowly increases the performance.

\begin{table}[h]
\centering
\footnotesize
\begin{tabular}{|c|c|c|}
\hline
& \textbf{w/o Clustering} & \textbf{With Clustering}  \\ \hline
\textbf{\#Topics}  & \textbf{F$_\text{mean}$}  & \textbf{F$_\text{mean}$}   \\ \hline
1 &  0.6244 & 0.5943 \\
3 & 0.6817 & 0.6322  \\
5 & 0.7134 & 0.6563  \\
7 & 0.7164 & 0.6703  \\
9 & 0.7151 & 0.6697 \\  
11 & 0.7305 & 0.6988 \\
13 & 0.7350 & 0.6964 \\
15 & 0.7370 & 0.7010 \\
17 & 0.7401 & 0.7034 \\ \hline
\end{tabular}
\caption{$F_1$ scores on the UKP ASPECT Corpus with increasing training set sizes (BERT model).}
\label{table_clustering_small_train_data}
\end{table}

\textbf{With Clustering.} We studied how the performance changes on the ASPECT corpus if we combine the similarity metric with agglomerative clustering (Table \ref{table_argument_clustering}). We notice that the performances drop by up to 7.64pp. Agglomerative clustering is a strict partitioning algorithm, i.e., each object belongs to exactly one cluster. However, an argument can address more than one aspect of a topic, therefore, arguments could belong to more than one cluster. Hence, strict partitioning clustering methods introduce a new source of errors.

We can estimate this source of error by evaluating the transitivity in our dataset. For a strict partitioning setup, if argument A $\sim$ B, and B $\sim$ C are similar, then A $\sim$ C are similar. This transitivity property is violated in 376 out of 1,714 (21.9\%) cases, indicating that strict partitioning is a suboptimal setup for the ASPECT dataset. This also explains why the human performance in the \textit{with clustering} setup is significantly lower than in the \textit{without clustering} setup.
As Table \ref{table_argument_clustering} shows, a better similarity metric must not necessarily lead to a better clustering performance with agglomerative clustering. Humans are better than the proposed BERT-model at estimating the pairwise similarity of arguments. However, when combined with a clustering method, the performances are on-par.

\section{Conclusion}
\label{sec:Conclusion}
Open-domain argument search, i.e. identifying and aggregating arguments for unseen topics, is a challenging research problem. The first challenge is to identify suitable arguments. 
Previous methods achieved low $F_1$-scores in a cross-topic scenario, e.g., \newcite{Stab2018b} achieved an $F_1$-score of .27 for identifying \textit{pro}-arguments. We could significantly improve this performance to .53 by using contextualized word embeddings. The main performance gain came from integrating topic information into the transformer network of BERT, which added 13pp compared to the setup without topic information.

The second challenge we addressed is to decide whether two arguments on the same topic are similar. Previous datasets on argument similarity used curated lists of arguments, which eliminates noise from the argument classification step. In this publication, we annotated similar argument pairs that came from an argument search engine. As the annotation showed, about 16\% of the pairs were noisy and did not address the target topic.   

Unsupervised methods on argument similarity showed rather low performance scores, confirming that fine-grained semantic nuances and not the lexical overlap determines the similarity between arguments. We were able to train a supervised similarity function based on the BERT transformer network that, even with little training data, significantly improved over unsupervised methods. 

While these results are very encouraging and stress the feasibility of open-domain argument search, our work also points to some weaknesses of the current methods and datasets.   
A good argument similarity function is only the first step towards argument clustering. We evaluated the agglomerative clustering algorithm in combination with our similarity function and identified it as a new source of errors. Arguments can address multiple aspects and therefore belong to multiple clusters, something that is not possible to model using partitional algorithms.
Future work should thus study the overlapping nature of argument clustering.
Further, more realistic datasets, that allow end-to-end evaluation, are required.

\section*{Acknowledgments}
The authors would like to sincerely thank Joy Mahapatra, who carried out the initial annotation study.
This work has been supported by the German Research Foundation through the German-Israeli Project Cooperation (DIP, grant DA 1600/1-1 and grant GU 798/17-1) and within the project ``Open Argument Mining'' (GU 798/25-1), associated with the Priority Program ``Robust Argumentation Machines (RATIO)'' (SPP-1999). It has been co-funded by the German Federal Ministry of Education and Research (BMBF) under the promotional references 01UG1816B (CEDIFOR) and 03VP02540 (ArgumenText).

\bibliography{acl2019}
\bibliographystyle{acl_natbib}

\appendix
\newpage
\section{Appendices} \label{sec_appendix}

 \subsection{UKP ASPECT Corpus: Amazon Mechanical Turk Guidelines and Inter-annotator Agreement}

 The annotations required for the UKP ASPECT Corpus were acquired via crowdsourcing on the Amazon Mechanical Turk platform.
 Workers participating in the study had to be located in the US, with more than 100 HITs approved and an overall acceptance rate of 90\% or higher. We paid them at the US federal minimum wage of \$7.25/hour. Workers also had to qualify for the study by passing a qualification test consisting of twelve test questions with argument pairs. Figure~\ref{fig:amt-guidelines} shows the instructions given to workers.

 \subsection{AFS Corpus: Detailed Results}

 Table \ref{tab:afs_corpus_full} shows the full results of the (un)supervised methods for the argument similarity calculation on the AFS dataset (all topics).

\begin{figure*}[!b]
\fbox{%
\small{
	\begin{minipage}{0.97\textwidth}
		Read each of the following sentence pairs and indicate whether they argue about the same aspect with respect to the given topic (given as ``Topic Name'' on top of the HIT). There are \textbf{four options}, of which one needs to be assigned to each pair of sentences (arguments). Please read the following for more details.
		\begin{itemize}
			\item \textbf{Different Topic/Can't decide}: Either one or both of the sentences belong to a topic different than the given one, or you can't understand one or both sentences. If you choose this option, you need to very briefly explain, why you chose it (e.g. ``The second sentence is not grammatical'', ``The first sentence is from a different topic'' etc.). For example, 
			
			Argument A: ``\textit{I do believe in the death penalty, tit for tat}''.
		
		 	Argument B: ``\textit{Marriage is already a civil right everyone has, so like anyone you have it too}''. 
				
			\item \textbf{No Similarity}: The two arguments belong to the same topic, but they don't show any similarity, i.e. they speak about completely different aspects of the topic. For example, 
			
			Argument A: ``\textit{If murder is wrong then so is the death penalty}''.
			
			Argument B: ``\textit{The death penalty is an inappropriate way to work against criminal activity}''.
				
			\item \textbf{Some Similarity}: The two arguments belong to the same topic, showing semantic similarity on a few aspects, but the central message is rather different, or one argument is way less specific than the other. For example, 
			
			Argument A: ``\textit{The death penalty should be applied only in very extreme cases, such as when someone commands genocide}''.
			
			Argument B: ``\textit{An eye for an eye: He who kills someone else should face capital punishment by the law}''. 
				
			\item \textbf{High Similarity}: The two arguments belong to the same topic, and they speak about the same aspect, e.g. using different words. For example, 
			Argument A: ``\textit{An ideal judiciary system would not sentence innocent people}''.
			
			Argument B: ``\textit{The notion that guiltless people may be sentenced is indeed a judicial system problem}''. 
			\end{itemize}
			
			Your rating should not be affected by whether the sentences attack (e.g. ``\textit{Animal testing is cruel and inhumane}'' for the topic ``\textit{Animal testing}'') or support (e.g. ``\textit{Animals do not have rights, therefore animal testing is fair}'' for the topic ``\textit{Animal testing}'') the topic, but only by the aspect they are using to support or attack the topic.
		
\end{minipage}
}}

    \caption{Amazon Mechanical Turk HIT Guidelines used in the annotation study for the Argument Aspect Similarity  Corpus.}
    \label{fig:amt-guidelines}
\end{figure*}

%\FloatBarrier

%%%%%%%%%%%%%%%%%%
%%%%%%% AFS %%%%%%
%%%%%%%%%%%%%%%%%%
\begin{table*}[t]
\centering
\footnotesize
\begin{tabular}{|l|c|c|c|c|c|c|c|c|}
\hline
& \multicolumn{2}{c|}{\textbf{Gun Control}}  & \multicolumn{2}{c|}{\textbf{Gay Marriage}}  & \multicolumn{2}{c|}{\textbf{Death Penalty}} & \multicolumn{2}{c|}{\textbf{Avg.}} \\
& $r$ & $\rho$ & $r$ & $\rho$ & $r$ & $\rho$ & $r$ & $\rho$ \\ \hline
 Human Performance & .6900 & - & .6000 & - & .7400 & - & .6767 & - \\
\hline
\multicolumn{9}{|l|}{\textbf{Unsupervised Methods}} \\
\hline
\model{Tf-Idf} & .6266 & .5528 & .4107 & .3778 & .3657 & .3589 & .4677 & .4298 \\
\model{InferSent - fastText} & .3376 & .3283 & .1012 & .1055 & .3168 & .2931 & .2519 & .2423  \\
\model{InferSent - GloVe}  & .3757 & .3707 & .1413 & .1435 & .2953 & .2847 & .2708 & .2663  \\
\model{GloVe Embeddings} & .4344 & .4485 & .2519 & .2741 & .2857 & .2973 & .3240 & .3400 \\
\model{ELMo Embeddings} & .3747 & .3654 & .1753 & .1709 & .2982 & .2663 & .2827 & .2675 \\
\model{BERT Embeddings} & .4575 & .4460 & .1960 & .1999 & .4082 & .4072 & .3539 & .3507 \\ \hline
\multicolumn{9}{|l|}{\textbf{Supervised Methods: Within-Topic Evaluation}}  \\ \hline
\model{SVR} \cite{MisraEW16}  & .7300 & - &  .5400 & - & .6300 & - &  .6333 & - \\ 
\model{BERT-base}  & .8323 & .8076 & .6255 & .6122 & .7847 & .7768 & .7475 & .7318 \\ 
\model{BERT-large}  & .7982 & .7592 & .6240 & .6137 & .7545 & .7149 & .7256 & .6959 \\ \hline 
\multicolumn{5}{|l|}{\textbf{Supervised Methods: Cross-Topic Evaluation}}  \\ \hline
\model{BERT-base}  & .6892 & .6689 & .4307 & .4236 & .6339 & .6245 & .5849 &  .5723 \\ 
\model{BERT-large} & .6895 &  .6749 & .5071 & .4866 & .6641 & .6486 & .6202 & .6034 \\ 
\hline
\end{tabular}
\caption{Pearson correlation $r$ and Spearman's rank correlation $\rho$ on the AFS dataset. Within-Topic Evaluation: 10-fold cross-validation. Cross-Topic Evaluation: System trained on two topics, evaluated on the third.}
\label{tab:afs_corpus_full}
\end{table*}

\end{document}